\documentclass{article}
\usepackage{spconf,amsmath,amssymb,graphicx,url}

\DeclareMathOperator{\logit}{logit}

\DeclareMathOperator{\erf}{erf}

\def\ND{\mathcal{N}}

\title{GENERATIVE MODELLING FOR UNSUPERVISED SCORE CALIBRATION}
\twoauthors
  {Niko Br\"ummer\sthanks{The experiments were done while attending the CLSP 2013 Speaker Recognition Workshop at Johns Hopkins University.}}
	{AGNITIO Research\\
	Somerset West, South Africa}
  {Daniel Garcia-Romero}
	{HLTCOE,
	Johns Hopkins University\\
	Baltimore, MD, USA}
\begin{document}
\ninept
\hyphenation{norm-DCF}
\maketitle
\begin{abstract}
Score calibration enables automatic speaker recognizers to make cost-effective accept / reject decisions. Traditional  calibration requires supervised data, which is an expensive resource. We propose a 2-component GMM for unsupervised calibration and demonstrate good performance relative to a supervised baseline on NIST SRE'10 and SRE'12. A Bayesian analysis demonstrates that the uncertainty associated with the unsupervised calibration parameter estimates is surprisingly small.   
\end{abstract}

\section{Introduction}
Automatic speaker recognizers map trials to scores. A \emph{trial} has two parts: some speech of a known speaker, and some of an unknown speaker. When the same speaker is present in both parts, we have a \emph{target} trial. When the speakers differ, we have a \emph{non-target} trial. The \emph{score} is a real number, which is expected to be larger (more positive) for target trials and smaller (more negative) for non-target trials. When a speaker recognizer is deployed in a new environment, which may differ from previously seen environments w.r.t.\ factors like language, demographics, vocal effort, noise level, microphone, transmission channel, duration, etc., the behaviour of the scores may change. Although the scores can still be expected to discriminate between targets and non-targets in the new environment, score distributions could change between environments.

If scores are to be used to make hard decisions, then we need to \emph{calibrate} the scores for the appropriate environment. The ideal calibration of a score, $s$, would be of the form:
\begin{align*}
s&\to\log\frac{P(s|\text{target},\text{environment})}{P(s|\text{non-target},\text{environment})}
\end{align*}
but of course, we are not given these score distributions. Our only resource would be some data collected from the new environment. In special cases (usually involving considerable expense), this data can be supervised, such that each trial is labelled as target or non-target. In a more realistic scenario however, all or most of this data would be \emph{unsupervised}. 

To date, most works on calibration have made use of supervised data. In this paper, we explore the problem of calibration where our only resource is a large database of completely unsupervised scores.

\section{Calibration model}
\def\hpr{\tau}
\def\model{\mathcal{M}}
\def\scores{\mathcal{S}}
\def\labels{\mathcal{L}}
In the supervised setting, score calibration can be viewed as a straight-forward, 2-class pattern recognition problem, for which both generative and discriminative solutions exist~\cite{Ramos-Castro:2005,art:brummer_csl_2006,art:Brummer_Fusion_TASLP_2007,DvL_IS13,NB_IS13}. For the  unsupervised case, we found the generative approach more convenient. Here we introduce the score model for supervised calibration, followed by a generalization to the unsupervised case.

\subsection{Supervised calibration model}
\label{sec:sup}
\def\EER{\text{EER}}
\def\cal{\mathcal{C}}
For the supervised case, we adopt the simple generative model of~\cite{DvL_IS13}. Denoting targets as $H_1$ and non-targets as $H_2$, we model a score $s\in\mathbb{R}$, with class-conditional Gaussian distributions:
\begin{align}
\label{eq:cal}
P(s|H_i,\cal) &= \ND(s|\mu_i,\sigma^2)
\end{align}
where $\mu_1,\mu_2$ are class-conditional means and $\sigma^2$ is the common within-class variance. We collectively refer to $\cal=(\mu_1,\mu_2,\sigma^2)$ as the \emph{calibration parameters}. This model gives an affine calibration transformation, from score to log-likelihood-ratio, of the form:
\begin{align}
\label{eq:plugin}
\log R(s|\cal) &= \log\frac{\ND(s|\mu_1,\sigma^2)}{\ND(s|\mu_2,\sigma^2)}=\frac{d'}{\sigma}s+\frac{\mu_2^2-\mu_1^2}{2\sigma^2}
\end{align}
where $d'$, the separation between targets and non-targets~\cite{book:Swets},
\begin{align}
d'&=\frac{\mu_1-\mu_2}{\sigma},
\end{align}
represents \emph{accuracy}, since the theoretical equal-error-rate is $\EER=\Phi(-\frac{d'}{2})$.\footnote{$\Phi$ is the \emph{normal cumulative density}, given in terms of the \emph{error-function} as $\Phi(x)=(1+\erf(x/\sqrt{2}))/2$.} At complete overlap, $d'=0$ and $\EER=0.5$. As $d'$ increases, the EER decreases.

We refer to $R(s|\cal)$ as the \emph{plug-in LR}, because it can be calculated only if $\cal$ is given. In reality, these parameters are not given, so they must be estimated before being plugged into~\eqref{eq:plugin}. When labelled calibration training scores are given, maximum-likelihood parameter estimates are straight-forward---see~\cite{DvL_IS13}, where this plug-in recipe is shown to give similar accuracy to logistic regression calibration.

\subsection{Unsupervised calibration model}
In the unsupervised case, we are given a collection of $T$ training scores, denoted $\scores=s_1,\ldots,s_T$, but we are \emph{not} given the corresponding class labels. Denoting these labels as $\labels=\ell_1,\ldots,\ell_T$, we treat them as hidden variables and our calibration model generalizes to a 2-component Gaussian mixture model (GMM), for which we need an additional mixture-weight parameter, $\pi_1$. Letting $\pi_2=1-\pi_1$, the \emph{GMM likelihood} is:
\begin{align}
\label{eq:gmm}
P(\scores|\model) = \prod_{t=1}^T \sum_{i=1}^2 \pi_i \ND(s_t|\mu_i,\sigma^2)
\end{align}
where $\model=(\cal,\pi_1)=(\mu_1,\mu_2,\sigma^2,\pi_1)$ are the GMM parameters.

Now consider a new \emph{test score}, $s'$, generated by the same model, and with associated hidden class label $\ell'\in\{H_1,H_2\}$. The conditional dependency structure of all the parameters and variables can be summarized in \emph{graphical model}~\cite{book:Bishop_PRML} notation as:
\begin{align}
\label{eq:gm}
\pi_1\to\labels\to\scores\gets\cal\to s'\gets\ell'\gets\pi_1'
\end{align}
where we have introduced an independent prior, $\pi_1'$, for $\ell'$. The importance of this diagram cannot be overstressed. It is used repeatedly below, to be able to remove irrelevant conditioning terms.\footnote{Observation at a node with convergent arrows induces dependency between variables linked via this node; when not observed, such nodes block dependency. Conversely, nodes with divergent or aligned arrows induce dependency when not observed; and block dependency when observed. A node is `observed' if it appears to the right of the $|$ in probability notation.}

Our end-goal will be to infer the value of $\ell'$, when $\scores$ and $s'$ are observed and $\pi_1'$ is given, but where $\pi_1,\labels,\cal$ are unknown nuisance variables. The result could be given as the posterior $P(\ell'|\scores,s',\pi_1')$, or equivalently\footnote{To see this, use~\eqref{eq:gm} to find: $\frac{P(H_1|\scores,s',\pi_1')}{P(H_2|\scores,s',\pi_1')}=\frac{\pi_1'}{1-\pi_1'}\times R(s'|\scores)$.}, as the \emph{predictive likelihood-ratio}:
\begin{align}
\label{eq:pred}
R(s'|\scores) &= \frac{P(s'|\ell'=H_1,\scores)}{P(s'|\ell'=H_2,\scores)}
\end{align} 
Our end-goal is the \emph{calibration} of $s'$ via the mapping:
\begin{align*}
s'&\to \log R(s'|\scores)
\end{align*}

\section{Inference}
\def\Ex#1#2{\big\langle#1\big\rangle_{#2}}
Here we discuss computational strategies to compute $R(s'|\scores)$. The computation involves summing over all of the hidden labels, $\labels$, and integrating out the parameters, $\model$. Unfortunately this cannot be done in closed form. If conjugate priors are used, the parameters can be integrated out in closed form, but this makes the labels dependent, so that summing them out requires an intractable sum over $2^T$ terms. Conversely, if you start with the labels, they can be summed out in closed form, using~\eqref{eq:gmm}, but then the parameters cannot be integrated out in closed form. For this work, we shall follow the latter route, because the parameter space is just 4-dimensional, allowing approximate integration in this space. 

The numerator and denominator of~\eqref{eq:pred} are obtained by marginalizing w.r.t.\ $\cal$ and simplifying by~\eqref{eq:gm}:
\begin{align}
\label{eq:canon}
R(s'|\scores) &= \frac{\Ex{P(s'|\cal,H_1)}{\scores}}{\Ex{P(s'|\cal,H_2)}{\scores}}
\end{align}
where $\langle\rangle_{\scores}$ denotes expectation w.r.t.\ the parameter posterior $P(\cal|\scores)$. We show below how to derive $P(\cal|\scores)$ from the \emph{Laplace approximation} for $P(\model|\scores)$.

Although we shall use~\eqref{eq:canon} in practice, we develop an interesting alternative form below that helps to theoretically illuminate the relationship between the plug-in and predictive LRs.

\subsection{Plug-in vs predictive LR}
\def\KL#1#2{D_\text{KL}(#1\|#2)}
The predictive likelihoods can be expanded by the product rule as:
\begin{align}
\label{eq:pre}
P(s'|S,H_i) &= P(s'|\cal,H_i)\times\frac{P(\cal|\scores)}{P(\cal|\scores,s',H_i)}
\end{align}
where a ratio of parameter posteriors modulates the plug-in likelihood. The numerator is conditioned on the unsupervised scores, while the denominator is conditioned on one additional supervised score, with assumed label $\ell'=H_i$. Although~\eqref{eq:pre} holds for any value of $\cal$ with non-zero posteriors, we find a more convenient form by taking logarithms and the expectation w.r.t.\ $P(\cal|\scores)$ on both sides:
\begin{align}
\label{eq:plh}
\log P(s'|S,H_i) &= \int \log P(s'|\cal,H_i) P(\cal|\scores) \,d\cal + D_i
\end{align}
where $D_i$ denotes KL-divergence between the posteriors:
\begin{align}
D_i &= \int P(\cal|\scores)\log\frac{P(\cal|\scores)}{P(\cal|\scores,s',H_i)}\,d\cal
\end{align} 
Using~\eqref{eq:plh} in~\eqref{eq:pred} gives the predictive log-LR as:
\begin{align}
\label{eq:pre3}
\log R(s'|S) &=  \Ex{\log R(s'|\cal)}{\scores} + D_1 - D_2
\end{align}
Notice that:
\begin{align}
\Ex{\log R(s'|\cal)}{\scores} &= s'\Ex{\frac{d'}{\sigma}}{\scores}+\Ex{\frac{\mu_2^2-\mu_1^2}{2\sigma^2}}{\scores}
\end{align}
which remains affine in $s'$, just like~\eqref{eq:plugin}. Moreover, if $P(\cal|\scores)$ has a \emph{sharp, dominant\footnote{By \emph{dominant}, we mean that the peak contains almost all of the probability mass in $P(\cal|\scores)$.} peak}, then $\Ex{\log R(s'|\cal)}{\scores}\approx\log R(s'|\hat\cal)$, where $\hat\cal$ is the mode of the dominant peak. Finally, if there are many scores in $\scores$, then a single additional score $s'$ that is similar to the scores in $\scores$, will result in small $D_i$, so that $\log R(s'|\scores)\approx\log R(s'|\hat\cal)$. Only if $\scores$ has very few scores, or $s'$ is very far away, will the $D_i$ cause significant non-linearity in $\log R(s'|\scores)$.

We already know that we have a large collection of unsupervised scores, but it remains to be demonstrated that $P(\cal|\scores)$ has a dominant peak, which we shall do below, via an experimental exploration of the likelihood. We shall also quantify the sharpness of that peak by using the \emph{Laplace approximation}.

\subsection{Laplace approximation}
\label{sec:LA}
\def\hess{\Lambda}
\def\mvec{\mathbf{m}}
\def\detm#1{\lvert#1\rvert}
\def\Cmat{\mathbf{C}}
The Laplace approximation (LA) is ideally suited to approximating sharply peaked, low-dimensional posteriors~\cite{book:Bishop_PRML,book:MacKay}. We have only 4 parameters and the likelihood is sharply peaked because we have lots of data. The only pitfall is that label swapping causes two identical peaks in the likelihood. We kill the unwanted peak by assigning a prior of the form $P(\model)\propto u(\mu_1-\mu_2)$, where $u$ is the unit step function. We do not need to specify the prior in any more detail, because any reasonable prior that we might want to assign here would be effectively constant relative to the sharply peaked likelihood.

Following the standard LA recipe to approximate the posterior $P(\model|\scores)$, we define:
\begin{align}
f(\model) &= \log P(\scores,\model)= \log [P(\scores|\model) P(\model)] 
\end{align}
which is computable by~\eqref{eq:gmm}. Notice $P(\model|\scores) \propto e^{f(\model)}$. Let $\hat\model$ be the dominant mode of $f$ and form a 2nd-order Taylor-series approximation here. The gradient at the mode is zero, but we need the Hessian (2nd derivative matrix), denoted $\hess$. This forms a multivariate Gaussian, approximate posterior:
\begin{align}
\label{eq:pmodel}
\tilde P(\model|\scores) &= \ND(\model|\hat\model,-\hess^{-1})
\end{align}
Since~\eqref{eq:canon} calls for $P(\cal|\scores)$, we still have to marginalize\footnote{Recall $\model=(\cal,\pi_1)$. By~\eqref{eq:gm}, $\pi_1$ and $\cal$ are dependent in $P(\cal,\pi_1|\scores)$, so we cannot just ignore $\pi_1$.}~\eqref{eq:pmodel} w.r.t.\ $\pi_1$. With the Gaussian approximation this is easy. Let $\Cmat=-\hess^{-1}$ denote the covariance of $\tilde P(\model|\scores)=\ND(\model|\hat\model,\Cmat)$, then the corresponding marginal $\tilde P(\cal|\scores)$ is also multivariate Gaussian, where the elements in $\hat\model$ and $\Cmat$ corresponding to $\pi_1$ have been removed~\cite{book:Bishop_PRML}. In summary, we get the 4-dimensional mode and Hessian, invert the Hessian and then discard one dimension.

\subsubsection{Model parametrization}
The LA is \emph{not} invariant to parametrization~\cite{book:MacKay}. Moreover, it is obvious that the true posterior for the parameters $\pi_1$ and $\sigma^2$ cannot be Gaussian. But, for a sharply peaked posterior, the parametrization is not that important and the behaviour far from the maximum is almost irrelevant. As long as $f(\model)$ is smooth enough so that a 2nd-order approximation is accurate close to the maximum, the posterior peak will be approximately Gaussian. If the likelihood magnitudes are large, then by the time the 2nd-order approximation becomes inaccurate, this inaccuracy becomes irrelevant because of the effect of the exponentiation. The reader is encouraged to consult Wikipedia,\footnote{\url{en.wikipedia.org/wiki/Laplace_approximation}} where this is graphically illustrated. Our experimental results below are reported for the parametrization $[\mu_1,\mu_2,\log(\sigma^2),\log(\pi_1)]$.

\section{Experiments}
We used two score corpora, \emph{DAC} and \emph{ABC}:\\

\noindent\textbf{The Domain Adaptation Challenge (DAC)}\footnote{\url{www.clsp.jhu.edu/workshops/archive/ws-13/}.} has telephone speech from the LDC's Switchboard and Mixer databases as well as NIST SRE'10~\cite{web:sre10}. It has three parts:

\textbf{Recognizer:} Switchboard was used to train the hyperparameters of an i-vector PLDA speaker recognizer~\cite{PLDA_LN}, which was then used to produce the scores below.

\textbf{Calibration:} About 7 million trials (single enrollment), from pre-SRE'10 Mixer corpora, provided the unsupervised scores, $\scores$, with target proportion about 4\% and (empirical) $\EER=2.38\%$.

\textbf{Evaluation:} About 400~000 trials, composed of pairs of segments from SRE'10, provided the \emph{test} scores, $s'$, with $\EER=5.54\%$.\\

\noindent\textbf{The ABC corpus} used a different speaker recognizer, applied to a different data set, drawn from the AGNITIO-BUT-CRIM submission to NIST SRE'12~\cite{ABC12,web:sre12}. The mixture of conditions was more diverse than for DAC, having telephone and microphone speech, variable number of enrollment segments, full and truncated test segments and varied noise levels.

\textbf{Recognizer:} Switchboard, Fisher and pre-SRE'12 Mixer was used to train an i-vector PLDA system. 

\textbf{Calibration:} About 42 million scores, $\scores$, from pre-SRE'12 Mixer. The target proportion is about 0.07\%, and $\EER=2.38\%$.

\textbf{Evaluation:} About 9 million test scores, $s'$, from SRE'12, with $\EER=3.25\%$.

\subsection{Exploration of likelihood}
The success of the whole venture depends critically on the behaviour of the GMM likelihood, $P(\scores|\model)$, given by~\eqref{eq:gmm}. If we are a-priori very uncertain about the proportion of targets in the unsupervised data, and also about the accuracy of the recognizer, it is not at all obvious whether there is enough information in the likelihood\footnote{The likelihood function, $P(\scores|\model)$, represents \emph{all} of the information that our chosen model can extract from $\scores$.} to be able to infer calibration parameters with a useful level of accuracy. Moreover, since our inference tools (plug-in and LA) both rely on finding likelihood optima, it is important to know whether the likelihood is plagued by local optima.\footnote{This true even for more sophisticated Bayesian tools, like variational Bayes and Gibbs sampling~\cite{book:Bishop_PRML}.}

To learn how the likelihood behaves, we did an exhaustive experimental exploration of the parameter space, $(\mu_1,\mu_2,\sigma^2,\pi_1)$. To facilitate visual representation, we used a 2-dimensional model representation, namely $(d',\log\frac{\pi_1}{1-\pi_1})$, which we plotted against log-likelihood, where the remaining two degrees of freedom were optimized.\footnote{Integrating them out using LA would also be feasible, but we found this unnecessary---when $d'$ and $\pi_1$ are fixed, there remains very little uncertainty about the scale and location. The constrained optimization was done with a bespoke EM algorithm, where the M-step had to make use of numerical optimization.} Recall from section~\ref{sec:sup} that $d'$ represents \emph{accuracy}, while $\pi_1$ represents \emph{target proportion}. If the likelihood has a single dominant peak for these two critical parameters, then there is hope that the calibration exercise will work. 

\begin{figure}[!t]
\centerline{
\includegraphics[width=8cm]{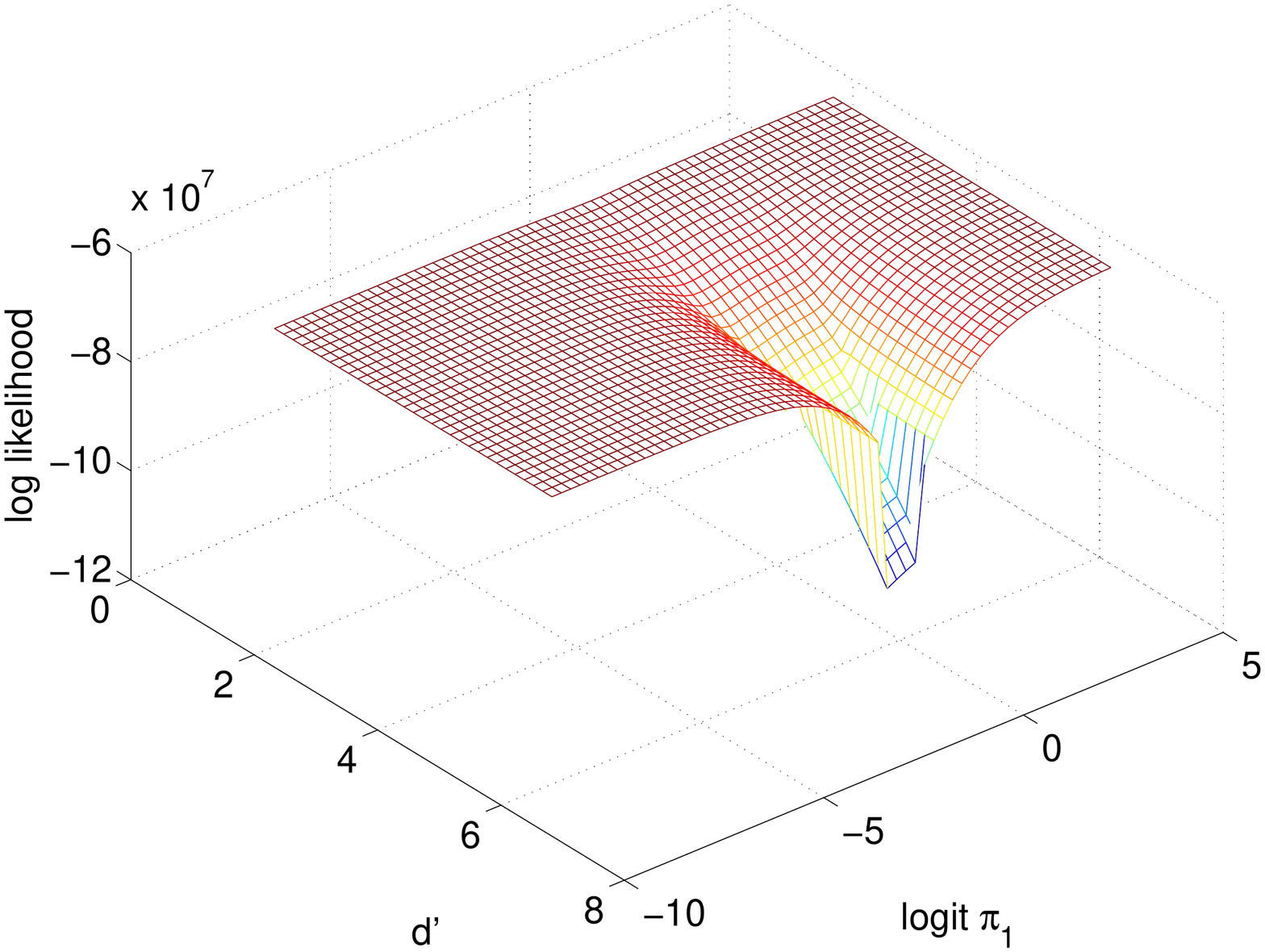}
}
\caption{$\log P(\scores|\model)$ for ABC}
\label{fig:LLH_mesh}
\end{figure}

\begin{figure}[!t]
\centerline{
\includegraphics[width=8cm]{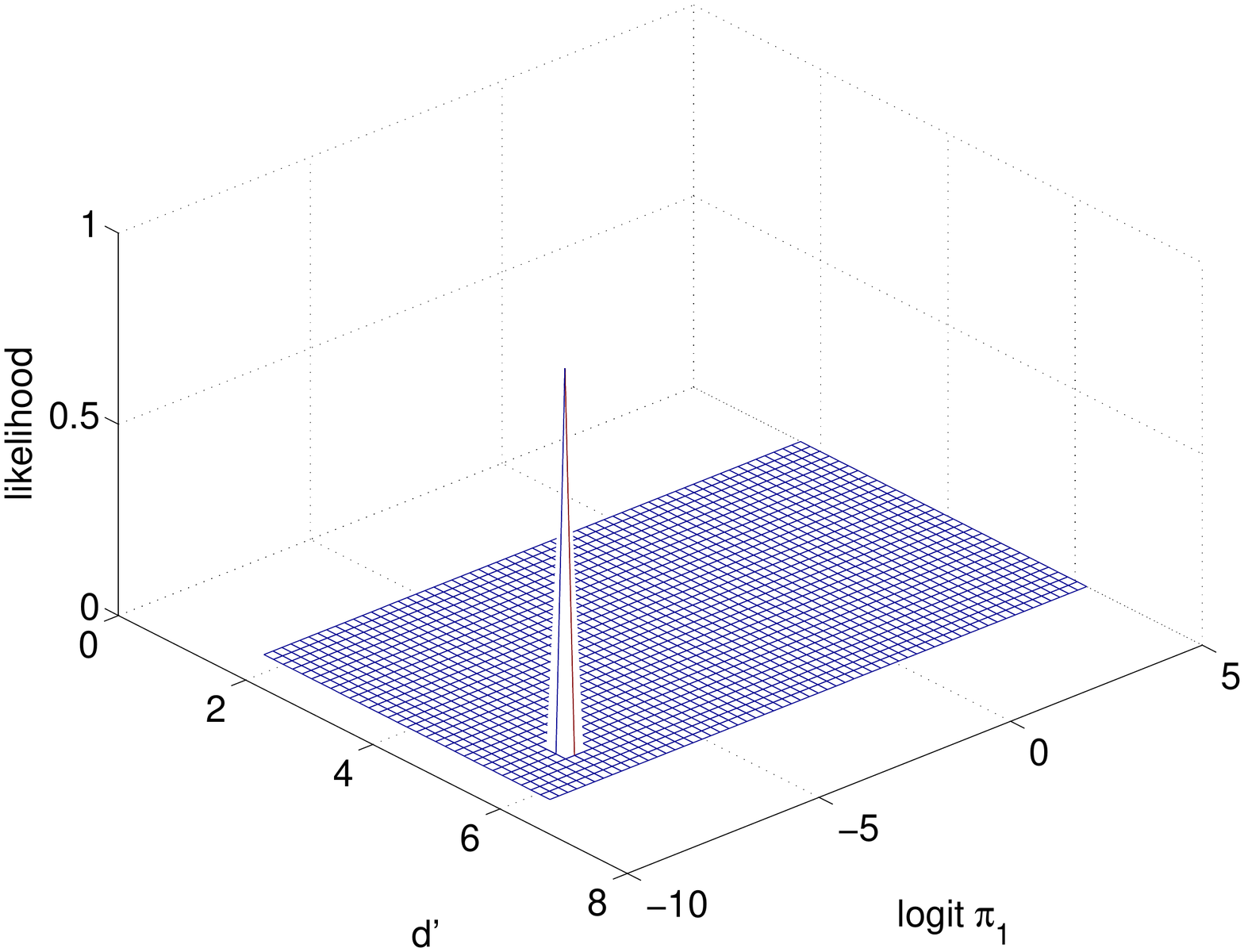}
}
\caption{normalized $P(\scores|\model)$ for ABC}
\label{fig:LH_mesh}
\end{figure}

We made such plots for the calibration parts of DAC and ABC. The results are similar. In figure~\ref{fig:LLH_mesh} we show the latter, which we believe is more challenging, because of the smaller target proportion. The log-likelihood is smooth as a function of $d'$, but is bi-modal as a function of $\pi_1$---a warning that initialization for the EM algorithm is important. Although the modes look flat, and similar, in the log-likelihood plot, the normalized\footnote{Subtract the maximum over the graph and then exponentiate~\cite{book:Sivia}.} likelihood plot of figure~\ref{fig:LH_mesh} reveals that there is just a \emph{single, sharp, dominant peak}, the location of which is given in table~\ref{table:params}.

\subsection{Analysis of sharpness}
To approximate $P(\model|\scores)$, we use the LA recipe of section~\ref{sec:LA}. The mode is found with the EM-algorithm. Complex-step differentiation~\cite{complexstep} and the Pearlmutter trick~\cite{Pearlmutter} are used for the Hessian. We find the \emph{error-bars} (posterior standard deviations~\cite{book:Sivia,book:MacKay}) for the parameters to be suprisingly small:\\

\centerline{
\begin{tabular}{|c|c|c|c|c|}
\hline
parametrization    &$\mu_1$ & $\mu_2$ & $\log(\sigma^2)$ & $\log(\pi_i)$ \\
\hline    
ABC error-bars & 0.0226 & 0.0004  & 0.0002           &  0.0071       \\
DAC error-bars & 0.1183 & 0.0192  & 0.0006           &  0.0022       \\
\hline
\end{tabular}
}

\subsection{Calibration experiments}
The sharpness of the parameter posterior shows there is no practical difference between the plug-in and predictive likelihood-ratios. We therefore proceed to report our final experimental results for maximum-likelihood plug-in calibration. We estimated the parameters on the calibration parts of DAC (7 million scores, 4\% targets) and ABC (42 million scores, 0.07\% targets). The performance of these calibrations was tested on the independent evaluation parts of those corpora, in terms of normalized Bayes error-rate, also known as normalized DCF~\cite{BOSARIS}:
\begin{align}
\text{normDCF}(\pi_1') &= \frac{\pi_1'P_\text{miss}(\pi_1')+(1-\pi_1)'P_\text{fa}(\pi_1')}{\min(\pi_1',1-\pi_1')}
\end{align}
where $P_\text{miss}(\pi_1')$ and $P_\text{fa}(\pi_1')$ are the empirical miss and false-alarm error-rates obtained when using the log-likelihood-ratios to make Bayes decisions at the theoretical threshold, $-\logit(\pi_1')$. The denominator is the Bayes error-rate for the default decision that always accepts, or always rejects, depending only on $\pi_1'$. Smaller values of normDCF are better, while a value of smaller than one shows the recognizer is doing better than the default decision. 

Table~\ref{table:calibration} reports normDCF for 4 different values of $\pi_1'$. For both databases, we compare the supervised recipe of~\cite{DvL_IS13} against the proposed unsupervised recipe. The supervised method used the \emph{same} data as the unsupervised recipe, except that the labels were supplied. We also report \emph{minDCF}, which uses an empirically optimized threshold at each operating point, where the optimization makes use of the evaluation labels. Finally, for DAC, we also report results for a supervised calibration trained on the \emph{mismatched}\footnote{Switchboard is at least a decade older than Mixer, during which time telephony changed dramatically~\cite{Mixer}.}, Switchboard data. The high error-rates for this case emphasizes the need for calibration on matched data. 

Surprisingly, for the DAC database, the unsupervised method does mostly better than the supervised one. This may be because of errors\footnote{Some Mixer subjects registered multiple times, with different PINs, thereby causing some target trials to be falsely labelled as non-targets.} in the labels supplied to the supervised method. 

In an effort to test whether our method holds up for very low target proportions, we noticed that we could go as far as removing \emph{all} trials labelled as targets, so that $\scores$ contained only trials labelled as non-targets. These entries in the table are marked as \emph{unsupervised*}. The fact that calibration still works in these cases can perhaps also be attributed to labelling errors.  

For additional insight, tables~\ref{table:params} and~\ref{table:calpar} compare the estimates of model and calibration parameters for the supervised and unsupervised cases.

\begin{table}[!htb]
\centerline{
\begin{tabular}{|l|c|c|c|c|}
\hline
$\pi_1'$           & 0.001  &  0.01 &  0.1   & 0.5   \\            
\hline
ABC supervised     & 0.32  &  0.22  &  0.13  &  0.08 \\
ABC unsupervised   & 0.33  &  0.24  &  0.16  &  0.11 \\
ABC unsupervised*  & 0.32  &  0.23  &  0.15  &  0.10 \\
ABC minDCF         & 0.31  &  0.21  &  0.12  &  0.06 \\
\hline
DAC mismatched     & 0.73  & 0.54   &  0.35  &  0.21 \\
DAC supervised     & 0.63  & 0.44   &  0.28  &  0.13 \\ 
DAC unsupervised   & 0.65  & 0.43   &  0.25  &  0.11 \\ 
DAC unsupervised*  & 0.65  & 0.43   &  0.24  &  0.12 \\  
DAC minDCF         & 0.63  & 0.42   &  0.24  &  0.11 \\  
\hline
\end{tabular}
}
\caption{Calibration performance in terms of normDCF.}
\label{table:calibration}
\end{table}

\begin{table}[!htb]
\centerline{
\begin{tabular}{|l|c|c|c|c|c|}
\hline
            & $\mu_1$ &  $\mu_2$ & $\sigma$ & $d'$ & $\pi_1$    \\            
\hline
ABC super   & 8.2  & -5.9  &  2.9  &  4.9 & 6.6e-4 \\
ABC unsup   & 9.9  & -5.9  &  2.9  &  5.5 & 5.6e-4 \\
ABC unsup*  & 9.6  & -5.9  &  2.9  &  5.4 & 5.1e-5 \\
\hline
DAC super   & 34.0 &-169.3 &  48.4 &  4.2 & 3.9E-2\\ 
DAC unsup   & 45.9 &-168.7 &  48.8 &  4.4 & 3.4E-2\\
DAC unsup*  & 72.3 &-169.3 &  48.0 &  5.0 & 1.4E-5\\
\hline
\end{tabular}
}
\caption{GMM parameter estimates}
\label{table:params}
\end{table}

\begin{table}[!htb]
\centerline{
\begin{tabular}{|l|c|c|}
\hline
            & scale &  offset \\            
\hline
ABC super     & 1.7   &   -2.0\\
ABC unsup   & 1.9   &   -3.8\\
ABC unsup*  & 1.9   &   -3.5\\
\hline
DAC super   &  0.087 &   5.9\\ 
DAC unsup   &  0.090 &   5.5\\
DAC unsup*  &  0.105 &   5.1\\
\hline
\end{tabular}
}
\caption{Calibration parameters}
\label{table:calpar}
\end{table}

\section{Conclusion}
The outcome of this work held two surprises for us. The first is that unsupervised calibration works at all. The second is that the missing labels contribute surprisingly little uncertainty to the parameter estimates. 

For future work on different data, we caution against blind application of the plug-in recipe. We feel that some Bayesian analysis similar to ours should also be done to illuminate the interaction between model and data.

\bibliographystyle{IEEEbib}
\bibliography{refs}

\end{document}